%% file: main.tex
\documentclass[conference]{IEEEtran}
\IEEEoverridecommandlockouts

\usepackage{cite}
\usepackage{amsmath,amssymb,amsfonts}
\usepackage{algorithmic}
\usepackage{graphicx}
\usepackage{textcomp}
\usepackage{xcolor}
\usepackage{booktabs}

\def\BibTeX{{\rm B\kern-.05em{\sc i\kern-.025em b}\kern-.08em
    T\kern-.1667em\lower.7ex\hbox{E}\kern-.125emX}}
\begin{document}

\title{Text-Guided Image Invariant Feature Learning for Robust Image Watermarking\\
}

\author{
    \IEEEauthorblockN{
    Muhammad Ahtesham\IEEEauthorrefmark{1}, 
    Xin Zhong\IEEEauthorrefmark{1}}\
    
    \smallskip
    \IEEEauthorblockA{
    \IEEEauthorrefmark{1}
    Department of Computer Science, University of Nebraska Omaha, Omaha, NE, USA
    \\ 
    mahtesham@unomaha.edu, xzhong@unomaha.edu
    }
}


\maketitle

\begin{abstract}
\input{section/abs}
\end{abstract}

\begin{IEEEkeywords}
Text-guided Invariant Feature Learning, Image Watermarking, Self-Supervised Learning.
\end{IEEEkeywords}

\vspace{-0.5em}
\section{Introduction}
\input{section/intro}

\vspace{-0.5em}
\section{Related Work}
\input{section/related}

\vspace{-0.5em}
\section{Proposed Method}
\input{section/method}

\vspace{-0.5em}
\section{Experiment and analysis}
\vspace{-0.5em}
\input{section/experiment}

\vspace{-0.5em}
\section{Conclusion}
\vspace{-0.5em}
\input{section/conclusion}

\newpage
\bibliographystyle{IEEEtran}
\bibliography{reference}

\end{document}

%% file: section/abs.tex
Ensuring robustness in image watermarking is crucial for and maintaining content integrity under diverse transformations. Recent self-supervised learning (SSL) approaches, such as DINO, have been leveraged for watermarking but primarily focus on general feature representation rather than explicitly learning invariant features. In this work, we propose a novel text-guided invariant feature learning framework for robust image watermarking. Our approach leverages CLIP's multimodal capabilities, using text embeddings as stable semantic anchors to enforce feature invariance under distortions. 
We evaluate the proposed method across multiple datasets, demonstrating superior robustness against various image transformations. 
Compared to state-of-the-art SSL methods, our model achieves higher cosine similarity in feature consistency tests and outperforms existing watermarking schemes in extraction accuracy under severe distortions. 
These results highlight the efficacy of our method in learning invariant representations tailored for robust deep learning-based watermarking. 

%% file: section/intro.tex
Image watermarking involves embedding hidden information within digital images to protect intellectual property and verify content authenticity. The embedded watermark is typically designed to remain imperceptible to human observers yet detectable and retrievable under various image transformations and distortions. As such, image watermarking plays an essential role in digital rights management, content authentication, and copyright protection.

Recent advancements in deep learning have significantly enhanced image watermarking techniques, resulting in methods that offer superior robustness and adaptability. Deep learning-based watermarking utilizes neural networks to learn rich and complex feature representations, enabling embedded watermarks to withstand common image transformations, including compression, cropping, rotation, scaling, and noise. Compared with traditional watermarking approaches, deep learning-based methods can adapt dynamically to varying embedding and extraction conditions, thus becoming the preferred solutions for modern multimedia applications~\cite{zhong2023brief, karki2024deep}.

A critical challenge in image watermarking is ensuring watermark robustness under complex, unpredictable image transformations. To achieve this, it is promising to explore invariant features that aim to extract robust image representations that remain stable despite significant distortions. Invariant features can ensure that the watermark remains reliably detectable, even after severe and diverse transformations. This robustness is crucial in practical applications, where images routinely undergo unpredictable modifications during storage, transmission, and processing.

State-of-the-art approaches in robust watermarking have leveraged self-supervised learning (SSL) methods, particularly models like DINO, to implicitly learn invariant representations. For instance, Fernandez et al.~\cite{fernandez2022watermarking} introduced a watermarking method using pre-trained DINO models to leverage a robust latent feature space. 
Such SSL-based techniques benefit from descriptive features learned through extensive general training, primarily optimized for tasks like classification, but do not specifically target robust invariant features essential for watermarking, potentially limiting their performance against image watermarking distortions.

In contrast, in this paper, we propose a novel text-guided learning framework specifically tailored for invariant feature learning for image watermarking. 
Our method utilizes textual embeddings as semantic anchors, guiding both original and distorted images toward shared, semantically meaningful representations. 
Specifically, the embedding of both original and distorted images is enforced to be semantically aligned with the text embedding, ensuring that invariant semantic content is preserved despite distortions. 
The primary contributions of our work are: (1) we introduce a novel method of using text as a semantic anchor to explicitly guide invariant feature learning in contrastive settings; and (2) we demonstrate significant improvements in robustness to complex and diverse image distortions within deep-learning-based watermarking scenarios. 

%% file: section/related.tex
\vspace{-0.5em}
\subsection{Invariant Feature Learning} 
\vspace{-0.5em}
Learning robust invariant features is a longstanding goal in computer vision, historically achieved through architectural innovations such as convolutional neural networks (CNNs), which incorporate slight invariances through operations like convolution and pooling~\cite{lecun1998cnn}. More recently, a paradigm shift toward data-driven approaches has emerged, enabling autonomous learning of invariances without predefined augmentations~\cite{autoaugment2018, zhang2021learning, immer2021invariance}. Techniques such as differentiable Kronecker-factored Laplace approximations~\cite{laplace1992, kfac2015} exemplify this trend, directly optimizing augmentation parameters during training rather than relying on manual augmentation strategies~\cite{immer2021invariance}. These methods adaptively identify features robust to transformations directly from the training data, enhancing feature stability across diverse conditions.

Self-supervised learning (SSL) methods, such as SimCLR~\cite{chen2020simple}, BYOL~\cite{grill2020bootstrap}, and DINO~\cite{caron2021emerging}, further contributed to this evolution by implicitly encouraging invariant representations through contrastive or teacher-student frameworks. Nevertheless, these SSL methods typically emphasize descriptive features optimized for downstream tasks such as classification or segmentation, rather than explicitly targeting robustness against image distortions relevant to watermarking scenarios. Recent developments in specialized tasks, including graph neural networks with frameworks like EQuAD~\cite{yao2024empowering}, have started to explicitly disentangle invariant features from spurious correlations. However, specialized invariant features for image watermarking, especially concerning specific robustness against complex distortions, remains under-explored.

Our proposed approach explicitly addresses this gap by directly training invariant representations anchored by textual descriptions through contrastive learning. Unlike existing SSL-based methods, where robustness emerges as a secondary byproduct of learning descriptive features, our model specifically optimizes representations to remain invariant to various transformations by using text embeddings as semantic anchors. This targeted invariance learning positions our method uniquely suitable for deep image watermarking applications.

\vspace{-0.5em}
\subsection{Deep Learning-based Image Watermarking.} 
\vspace{-0.5em}
Recent advancements in deep learning-based image watermarking techniques have substantially improved watermark robustness against common image distortions. Broadly, these methods fall into three categories~\cite{zhong2023brief}: (1) Embedder-Extractor Joint Training methods, which jointly train encoder-decoder networks end-to-end, integrating explicit noise layers to ensure watermark robustness; (2) Hybrid Approaches, combining jointly trained watermarking networks with pre-trained models to balance embedding flexibility and feature robustness; and (3) Deep Networks as Feature Transformation methods, which utilize pre-trained neural networks to extract fixed robust representations, enabling watermark embedding in stable latent spaces. Our work closely aligns with the third category by leveraging neural network-derived invariant features as a robust foundation for watermark embedding.

Within this paradigm, Vukotic et al.\cite{vukotic2020classification} utilized pre-trained CNNs to guide gradient-based embedding, balancing watermark detectability and image perceptual quality. Fernandez et al.\cite{fernandez2022watermarking} extended this approach by embedding multi-bit watermarks within invariant spaces derived from self-supervised frameworks such as DINO. This approach benefits significantly from the pre-existing robustness inherent in SSL-based learned features, albeit without explicitly training these features for robustness in watermarking contexts.

Our method expands upon this foundational approach, introducing a novel text-guided contrastive learning strategy specifically tailored to directly learn invariant representations. By explicitly anchoring image features to textual semantic embeddings, our method optimizes for robustness against targeted image transformations, making it particularly effective for watermark embedding tasks requiring resilience to diverse and unpredictable distortions.

%% file: section/method.tex
\begin{figure*}[!htb]
    \centering
    \vspace{-1.5em}
    \includegraphics[width=0.725\linewidth]{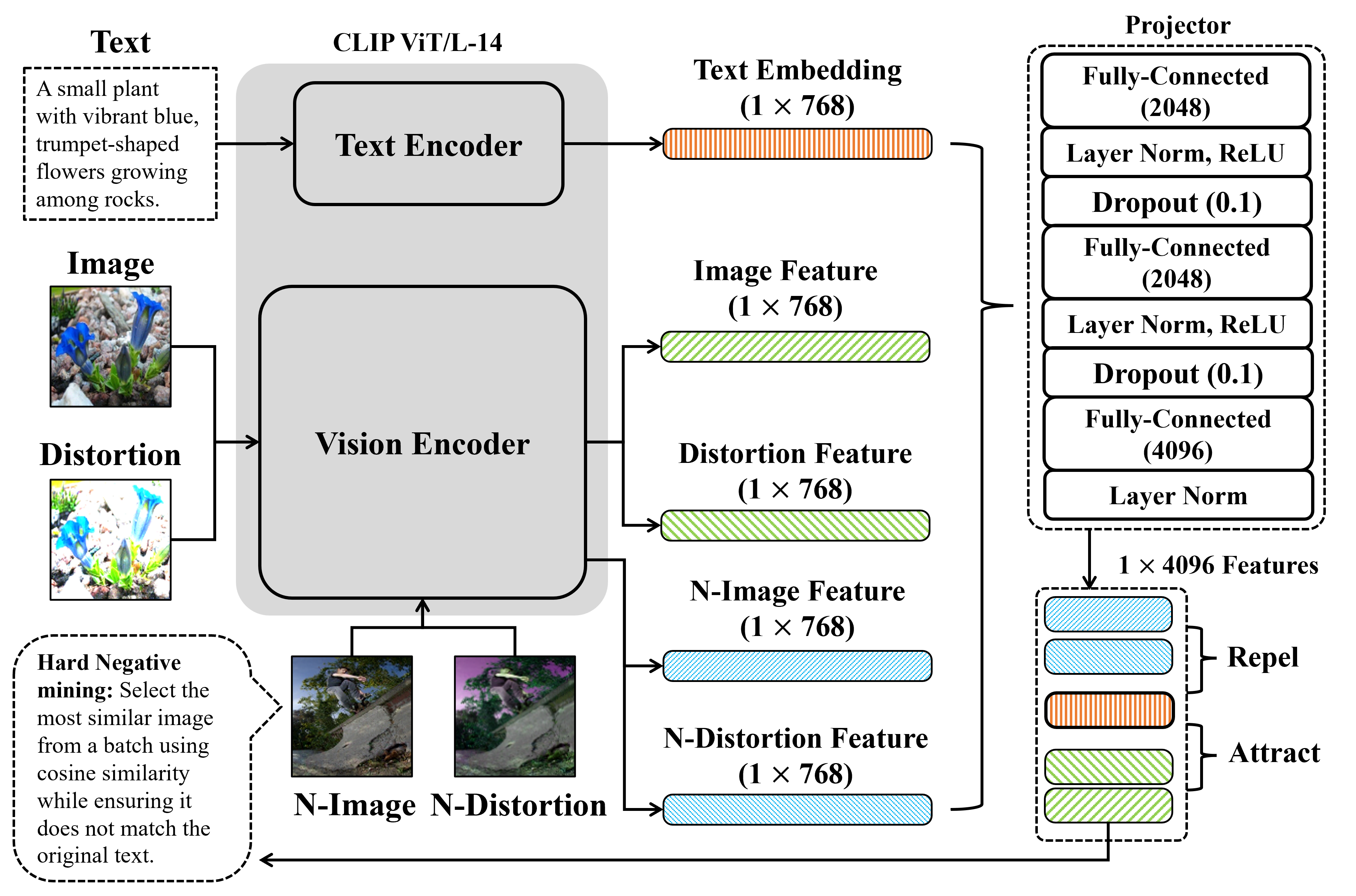}
    \vspace{-1.0em}
    \caption{The proposed text-guided invariant feature learning.}
    \label{fig:inv_feature}
    \vspace{-1.5em}
\end{figure*}

This section presents the proposed text-guided learning framework for invariant features in image watermarking. 
Our method leverages a multimodal approach by using the CLIP~\cite{radford2021learning}, which extracts feature representations from both images and their corresponding textual descriptions. As shown in Fig.~\ref{fig:inv_feature}, the framework consists of the following main components: the CLIP feature extractor (text encoder and vision encoder), and a custom projector.

Given an image and its distorted counterpart, the vision encoder extracts separate feature embeddings for both versions, ensuring that their representations align semantically with the corresponding text embedding obtained from the text encoder. To enhance invariance learning, we employ hard negative mining by selecting visually similar images from the dataset that do not match the given textual description. This ensures that the model learns to distinguish between semantically relevant and irrelevant features. 
The extracted text and image embeddings are then projected into a higher-dimensional space using a custom projector, which consists of fully connected layers with layer normalization and dropout for stability. Finally, a contrastive loss is applied to maximize similarity between corresponding image-text pairs while repelling hard negatives, thereby enforcing invariance across different distortions while preserving semantic alignment.

Unlike traditional image-based contrastive learning, which relies solely on visual similarity and may struggle with distortions that alter low-level image features, our approach leverages text embeddings as stable semantic anchors. Text descriptions capture high-level semantics that remain invariant under transformations, ensuring that both the original and distorted images align with the same semantic meaning. This explicit semantic grounding allows our model to learn robust invariant features that are invariant to diverse image distortions while maintaining meaningful representation consistency.

\vspace{-0.5em}
\subsection{Feature Extractor}
\vspace{-0.5em}
\noindent To handle both text and image modalities, we employ CLIP~\cite{radford2021learning} as our feature extractor, leveraging its dual-encoder architecture: a Vision Transformer (ViT-L/14) for images and a Text Transformer for textual descriptions. This setup aligns images and text within a shared feature space, making it highly effective for text-guided feature learning.

Among CLIP’s variants (ViT-B/32, ViT-B/16, ViT-L/14, ViT-H/14), we choose ViT-L/14 for its balance between efficiency and performance. It processes images by resizing them to 224×224 pixels, dividing them into 14×14 patches (16×16 pixels each), and projecting them into a 1024-dimensional embedding space, followed by 24 transformer layers with 16 self-attention heads. The final output is a 768-dimensional image feature vector. 

The text encoder tokenizes input text and processes it through a transformer-based architecture, producing a 768-dimensional text feature vector that captures semantic meaning. We extract these raw feature vectors and refine them through a custom projection head, optimizing their alignment for contrastive learning. ViT-L/14’s smaller patch size helps retain spatial details in images, while its textual encoding enhances semantic understanding, ensuring robust invariant feature extraction.

\vspace{-0.5em}
\subsection{Projector}
\vspace{-0.25em}
\noindent To enhance feature representations, we design a high-dimensional projector that maps 768-dimensional CLIP features to a 4096-dimensional space. Unlike conventional projection heads that reduce dimensionality, our approach preserves fine-grained information and improves feature separation for contrastive learning. Expanding the feature space enables the model to capture complex relationships while minimizing information loss.

The projector consists of a fully connected layer expanding features from 768 to 2048 dimensions, followed by Layer Normalization (LayerNorm) for activation stability, ReLU for non-linearity, and dropout (0.1) for regularization. The hidden layer maintains 2048 dimensions with the same transformations. The final output layer projects features to 4096 dimensions, maximizing encoding capacity. L2 normalization ensures unit-norm constraints, reducing redundancy and improving feature decorrelation. 
We use LayerNorm over BatchNorm due to its sample-wise normalization, ensuring stable activations across varying batch sizes—crucial for multimodal learning. We find that this combination of LayerNorm, non-linearity, dropout, and L2 normalization enhances feature stability, expressiveness, and discrimination.

\vspace{-0.5em}
\subsection{Training} 
\vspace{-0.5em} 

\noindent We train our model using the Flickr8k dataset~\cite{hodosh2013framing}, where each of approximately 8000 images is paired with five unique captions. 
To construct training pairs, we associate each image with its corresponding caption, resulting in roughly 40,000 image-caption pairs. 
To further enhance diversity, we apply a series of random distortions through an augmentation pipeline, as summarized in Table~\ref{tab:probabilities} and illustrated in Figure~\ref{fig:inv_feature}. 

Each distortion is applied with a predefined strength and an associated probability determining whether it is added or not. 
Each image is processed five times, once per caption, with a unique combination of distortions applied randomly. This ensures that each image generates multiple degraded variants, enriching the training signal for learning invariant representations. During training, we freeze the pre-trained CLIP model and optimize only the projector. 
This prevents modifications to CLIP’s original feature space while enabling our projector to refine representations for invariance. 
\noindent Table~\ref{tab:training_testing} presents the distortions applied in training and testing. 

\vspace{-0.5em}
\begin{table}[h]
    \centering
    \renewcommand{\arraystretch}{1.2}
    \small 
    \setlength{\tabcolsep}{4pt} 
    \resizebox{0.8\linewidth}{!}{ 
    \begin{tabular}{l p{3.5cm} c} 
        \toprule
        \textbf{Distortion Type} & \textbf{Parameters} & \textbf{Probability} \\
        \midrule
        Random Rotation & $\pm 30^\circ$ & 100\% \\
        Horizontal Flip & - & 100\% \\
        Color Jitter & Brightness=0.2, Contrast=0.2, Saturation=0.2, Hue=0.1 & 100\% \\
        Random Resized Crop & Scale=(0.8,1.0) to 224×224 & 100\% \\
        Minimal Noise & AddNoise (std=5) & 50\% \\
        Gaussian Blur & Kernel=5, Sigma=(0.1,2.0) & 50\% \\
        Strong Noise & AddNoise (std=25) & 30\% \\
        \bottomrule
    \end{tabular}
    }
    \vspace{0.5em}
    \caption{Distortion types, parameters, and probabilities.}
    \label{tab:probabilities}
    \vspace{-1.75em}
\end{table}

\begin{table}[h]
    \centering
    \renewcommand{\arraystretch}{1.2}
    \begin{tabular}{l c c}
        \toprule
        \textbf{Distortion Type} & \textbf{Training} & \textbf{Testing} \\
        \midrule
        Random Rotation & \textcolor{green}{\checkmark} & \textcolor{green}{\checkmark} \\
        Horizontal Flip & \textcolor{green}{\checkmark} & \textcolor{red}{$\times$} \\
        Color Jitter & \textcolor{green}{\checkmark} & \textcolor{green}{\checkmark} \\
        Random Resized Crop & \textcolor{red}{$\times$} & \textcolor{green}{\checkmark} \\
        Minimal Noise & \textcolor{green}{\checkmark} & \textcolor{red}{$\times$} \\
        Gaussian Blur & \textcolor{green}{\checkmark} & \textcolor{red}{$\times$} \\
        JPEG Compression & \textcolor{red}{$\times$} & \textcolor{green}{\checkmark} \\
        Solarization & \textcolor{red}{$\times$} & \textcolor{green}{\checkmark} \\
        Saturation & \textcolor{red}{$\times$} & \textcolor{green}{\checkmark} \\
        \bottomrule
    \end{tabular}
    \vspace{0.5em}
    \caption{Distortion usage in training and testing. 
    }
    \label{tab:training_testing}
    \vspace{-2.5em}
\end{table}

The training process is guided by a combination of contrastive loss and decorrelation loss. The contrastive loss encourages high cosine similarity between positive image-text pairs while penalizing negative pairs. Additionally, decorrelation loss is introduced to enhance feature diversity by reducing redundancy in the learned representations. The overall loss function is formulated as:
\vspace{-0.25em}
\begin{equation}
\mathcal{L}_{\text{total}} = \mathcal{L}_{\text{pos}} + \mathcal{L}_{\text{neg}} + \lambda_{\text{decorr}} \, \mathcal{L}_{\text{decorr}},
\end{equation}
\vspace{-1.5em}

\noindent where \( \lambda_{\text{decorr}} \) controls the decorrelation loss contribution.

Cosine similarity between two embeddings \( z_i \) and \( z_j \) is computed as:
\(
\text{sim}(z_i, z_j) = (z_i \cdot z_j) / (\|z_i\| \, \|z_j\|).
\)
We define cosine similarity for various feature pairs:
\(
\text{sim}_{\text{img,text}} = \text{sim}(z_{\text{img}}, z_{\text{text}}), 
\text{sim}_{\text{dist,text}} = \text{sim}(z_{\text{dist}}, z_{\text{text}}), 
\text{sim}_{\text{neg-img,text}} = \text{sim}(z_{\text{neg-img}}, z_{\text{text}}), 
\text{sim}_{\text{neg-dist,text}} = \text{sim}(z_{\text{neg-dist}}, z_{\text{text}}).
\)
To encourage high similarity for positive image-text pairs, the positive pair loss is defined as:
\begin{multline}
\mathcal{L}_{\text{pos}} = -\frac{1}{N} \sum_{i=1}^{N} \Biggl[
\log \frac{\exp\bigl(\text{sim}_{\text{img,text}}\bigr)}
{\exp\bigl(\text{sim}_{\text{img,text}}\bigr) + \exp\bigl(\text{sim}_{\text{neg-img,text}}\bigr)} \\
+ \,
\log \frac{\exp\bigl(\text{sim}_{\text{dist,text}}\bigr)}
{\exp\bigl(\text{sim}_{\text{dist,text}}\bigr) + \exp\bigl(\text{sim}_{\text{neg-dist,text}}\bigr)}
\Biggr].
\end{multline}

\noindent A margin-based negative loss is applied to ensure that negative pairs remain sufficiently distinct:
\begin{multline}
\mathcal{L}_{\text{neg}} = \frac{1}{N} \sum_{i=1}^{N} \Biggl[
\max\Bigl(0,\, \text{sim}_{\text{neg-img,text}} - \text{sim}_{\text{img,text}} + m\Bigr) \\
+ \, \max\Bigl(0,\, \text{sim}_{\text{neg-dist,text}} - \text{sim}_{\text{dist,text}} + m\Bigr)
\Biggr],
\end{multline}
where \( m \) is a predefined margin. 
Negative samples are selected within each batch by computing cosine similarities among image features. If the maximum similarity between a sample and any other exceeds a threshold \( \tau \), that sample is chosen as a hard negative; otherwise, a random negative is selected. This negative mining strategy ensures that the negatives used in the contrastive loss are challenging, thereby enhancing the model’s discriminative capability.

To enhance diverse feature representations for different images—even if they share similar visual appearances but differ in semantic meaning—we incorporate a decorrelation loss that centers the feature vectors and computes their covariance matrix \( C \). Specifically, given a feature tensor \( Z \in \mathbb{R}^{N \times d} \) (with \( N \) as the batch size and \( d \) as the feature dimension), we first center the features by subtracting the mean,
\(
\tilde{Z} = Z - \frac{1}{N}\sum_{i=1}^{N} z_i,
\)
and then compute the covariance matrix as
\(
C = \frac{1}{N}\tilde{Z}^\top \tilde{Z}.
\)
Redundancy is reduced by penalizing the squared off-diagonal elements of \( C \) via
\vspace{-0.25em}
\begin{equation}
\mathcal{L}_{\text{decorr}}(Z) = \sum_{i\neq j}\left(C_{ij}\right)^2 = \| C - \operatorname{diag}(C) \|_F^2.
\end{equation}
The decorrelation loss is applied to image, distribution, and text features as
\(
\mathcal{L}_{\text{decorr}} = \text{decorrelation\_loss}(z_{\text{img}}) + \text{decorrelation\_loss}(z_{\text{dist}}) + \text{decorrelation\_loss}(z_{\text{text}}).
\)



\vspace{-0.5em}
\subsection{Multi-bit Image Watermarking}
\vspace{-0.25em}
\label{sec: wm_method}

To embed a watermark into an image while preserving perceptual quality, we adopt a strategy based on data augmentation and back-propagation~\cite{vukotic2020classification, fernandez2022watermarking}, leveraging existing deep networks as feature transformations. 
Given an original image \( I_o \), our goal is to produce a marked image \( I_w \) that remains visually similar to \( I_o \) while ensuring that its feature representation aligns within a designated region \( D \) in the feature space. This region \( D \) is determined based on a trainable secret key and, in the multi-bit setting, the message to be embedded. The watermark embedding is optimized using the following objective function:

\vspace{-1.5em}
\begin{equation}
L(I_w, I_o, t) = \lambda L_w (\phi(\text{Tr}(I_w, t))) + L_i (I_w, I_o),
\end{equation}
\vspace{-1.5em}

\noindent where \( L_w \) enforces that the transformed image features \( \phi(\text{Tr}(I_w, t)) \) remain within the desired embedding space \( D \), while \( L_i \) ensures minimal perceptual distortion relative to \( I_o \).

The optimization is performed iteratively via stochastic gradient descent. In each iteration, a random transformation \( t \sim T \) (e.g., rotation, cropping, or blur) is applied to \( I_w \). The feature extractor \( \phi \) (i.e., our proposed invariant feature extractor) computes the image features, and the loss gradient updates \( I_w \), ensuring that the image remains perceptually similar to \( I_o \) while embedding the watermark effectively. 
A binary message \( m = \{b_1, b_2, ..., b_k\} \) is embedded by modulating the dot product between the image feature \( \phi(I_w) \) and a set of secret key vectors \( a_1, ..., a_k \). The embedding process minimizes the following hinge loss:

\vspace{-1.0em}
\begin{equation}
L_w(x) = \frac{1}{k} \sum_{i=1}^{k} \max(0, \mu - (\phi(I_w)^T a_i) \cdot b_i),
\end{equation}
\vspace{-1.0em}

\noindent where \( \mu \) is a margin ensuring that the extracted message remains robust to distortions. The gradient update is applied to the image \( I_w \), modulating its pixel values such that the dot product between its feature representation \( \phi(I_w) \) and the key vectors \( a_i \) aligns with the intended binary message \( m \). This optimization process ensures that the embedded watermark is both robust and imperceptible while preserving the original image quality. 
During extraction, the embedded watermark is retrieved by computing
\(
D(I_w) = \left[ \text{sign} (\phi(I_w)^T a_1), ..., \text{sign} (\phi(I_w)^T a_k) \right].
\)

Adopting this deep neural network feature extractor strategy enables us to leverage our proposed text-guided invariant feature extractor, resulting in high watermarking robustness.

%% file: section/experiment.tex
This section presents the experimental results and analysis of the proposed text-guided image invariant feature learning for robust image watermarking. 
In Section~\ref{inv_f_ana}, we evaluate the learned invariant features across different datasets and assess their tolerance to distortions. Section~\ref{inv_compare} compares our method against state-of-the-art SSL models across multiple datasets, demonstrating the robustness of our text-guided invariance approach. 
Finally, in Section~\ref{robust_watermarking}, we validate the effectiveness of the proposed invariant features for robust image watermarking under various distortions.

\vspace{-0.5em}
\subsection{Invariance Analysis of the Proposed Feature}
\label{inv_f_ana}
\begin{figure}[h]
    \centering
    \includegraphics[width=0.49\linewidth]{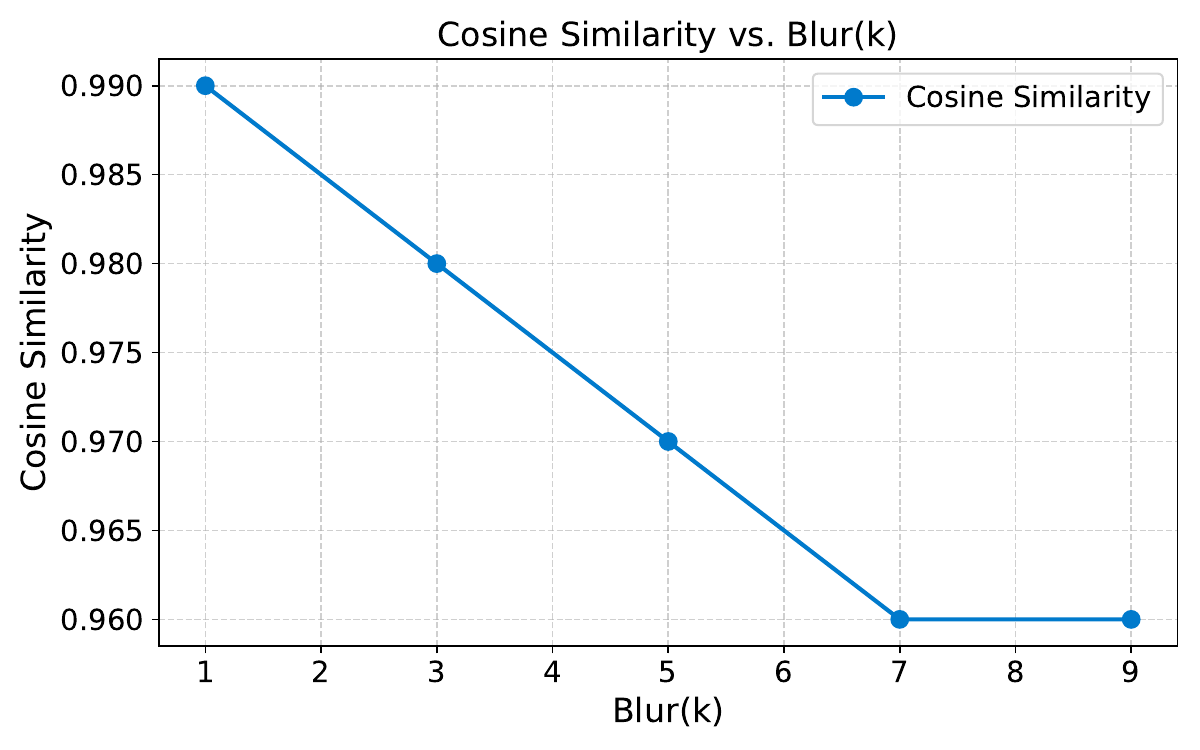}
    \includegraphics[width=0.49\linewidth]{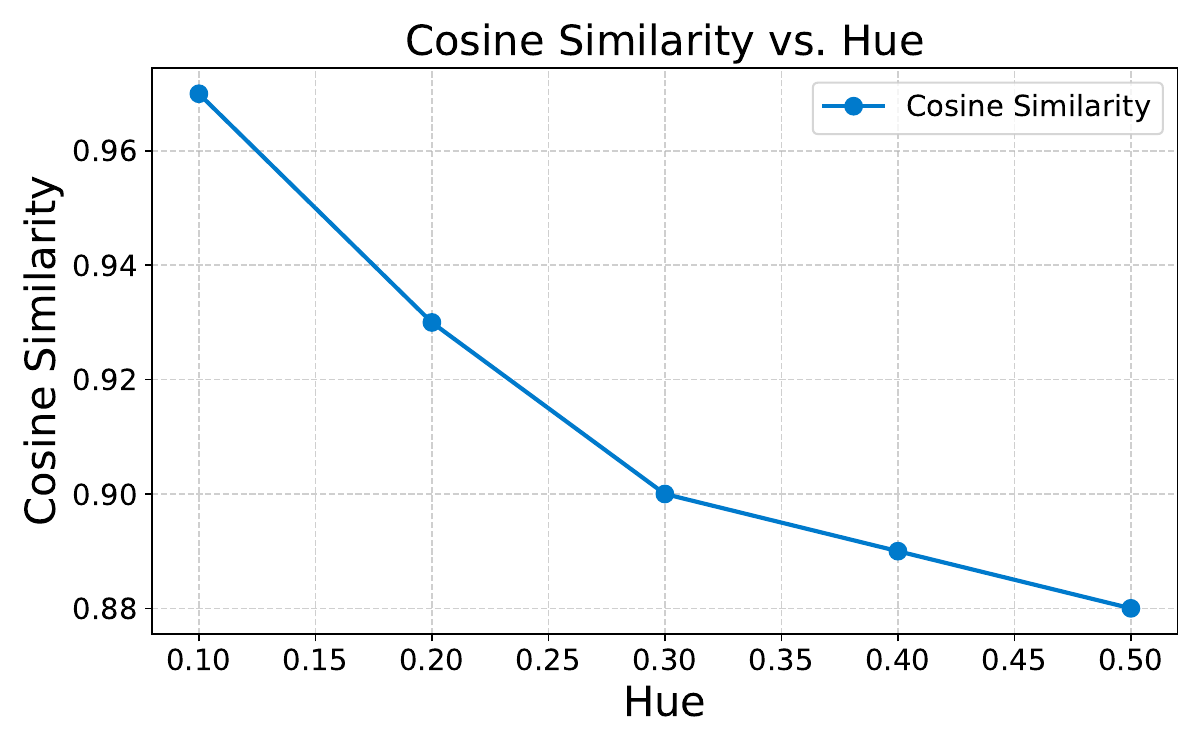}
    \vspace{-2.0em}
    \caption{Example tolerance of the proposed on different noise levels.}
    \label{fig:tolerance}
    \vspace{-1.0em}
\end{figure}

\vspace{-0.25em}
We conduct a comprehensive analysis of the proposed invariant features under various distortions across multiple datasets. 
In the first experiment, we test our approach to the Flickr8K dataset~\cite{hodosh2013framing}, using 36,000 images for training and 4,000 images for testing. 
During evaluation, we assess the model’s tolerance to distortions by using it as a feature extractor and computing cosine similarities between the features of original and distorted images. We systematically increase the distortion strength and observe the corresponding decrease in cosine similarity, measuring the extent to which the invariant features remain stable under transformations such as rotation, Gaussian blur, and hue shifts. Some results are presented in Fig.~\ref{fig:tolerance}. 
We can observe that even as the strength of hue or blur increases, the cosine similarity remains high and the proposed method shows high robustness by enhanced tolerance range against distortions.

\vspace{-1.0em}

To verify the robustness of the proposed features, we further evaluated our approach on three additional datasets—Flickr30k~\cite{plummer2015flickr30k}, OxfordPet III~\cite{parkhi2012cats}, and STL10~\cite{article}—to assess under various distortions. These datasets were selected for their diversity in content, including natural scenes, objects, and animals, ensuring a comprehensive evaluation of our model's performance. For each dataset, we extracted a subset of 1,000 images and computed the cosine similarity between the feature representations of original and distorted images under transformations such as rotation, brightness, contrast, hue, and blur.

\begin{table}[h]
\vspace{-1.0em}
\centering
\renewcommand{\arraystretch}{1.2} 
\begin{tabular}{lccc} \hline
 & \textbf{Flickr30k} & \textbf{Oxford Pet III} & \textbf{STL10} \\ \hline
(Rotation) 10   & 0.95 & 0.96 & 0.91 \\
(Rotation) 30   & 0.94 & 0.95 & 0.88 \\
(Brightness) 0.3 & 0.98 & 0.98 & 0.96 \\
(Brightness) 0.5 & 0.98 & 0.98 & 0.96 \\
(Contrast) 0.3   & 0.98 & 0.98 & 0.96 \\
(Contrast) 0.5   & 0.98 & 0.98 & 0.96 \\
(Hue) 0.1     & 0.97 & 0.97 & 0.95 \\
(Hue) 0.3     & 0.91 & 0.88 & 0.88 \\
(Blur) \(k=3\)   & 0.97 & 0.96 & 0.90 \\
(Blur) \(k=7\)   & 0.96 & 0.95 & 0.85 \\ \hline
\end{tabular}
\vspace{0.5em} 
\caption{Cosine Similarity Results on Different Datasets}
\label{tab:cosine_similarity_datasets}
\vspace{-2.5em}
\end{table}

These distortions were selected as they represent challenging conditions in image watermarking, where preserving invariant features is critical for robustness. For Flickr30k and OxfordPet III, cosine similarities remain consistently high across distortions, whereas STL10 exhibits slightly lower values, likely due to its distinct image characteristics. The results, summarized in table~\ref{tab:cosine_similarity_datasets}, demonstrate the effectiveness and stability of our approach across diverse datasets.

\vspace{-0.5em}
\subsection{Invariance Comparison}
\label{inv_compare}
\vspace{-0.25em}

Firstly, we compare the proposed invariant features against two SSL approaches: SimCLR~\cite{chen2020simple }, which utilizes contrastive learning to maximize agreement between differently augmented views, and DINO~\cite{oquab2024dinov2}, which employs self-distillation to learn robust representations without requiring labeled data. DINO has also been explored in other image watermarking schemes~\cite{fernandez2022watermarking} due to its ability to learn invariant feature spaces. 

For evaluation, we use subsets from the Oxford-IIIT Pet dataset (111 classes) ~\cite{parkhi2012cats} and the COCO dataset ~\cite{lin2014microsoft}. All images are resized to \(224 \times 224\), and a series of strong distortions are applied to rigorously assess the resilience of the learned features. The applied distortions include rotation by \(30^\circ\) and \(70^\circ\), cropping with removal of 2\% and 6\% of the image, and salt-and-pepper noise with probabilities of 0.05 and 0.10. Additionally, perspective distortion is tested under two configurations: one with a distortion scale of \(D=0.7\) and probability \(p=0.7\), and another with extreme settings of \(D=1.0\) and \(p=1.0\), where \(D\) controls the maximum magnitude of the distortion and \(p\) determines the likelihood of applying the transformation. 
Tables~\ref{tab:oxford_compare} and~\ref{tab:coco_compare} present the cosine similarity between the feature representations of original and distorted images for each method.

\vspace{-0.75em}
\begin{table}[ht]
\centering
\renewcommand{\arraystretch}{1.2} 

\begin{tabular}{lccc} 
\hline
\textbf{Distortion} & \textbf{SimCLR} & \textbf{DINO} & \textbf{Our Model} \\ \hline
(Rotation) \(30^\circ\)   & 0.90 & 0.95 & 0.95 \\
(Rotation) \(70^\circ\)   & 0.84 & 0.88 & 0.91 \\
(Crop) 2\%               & 0.54 & 0.48 & 0.68 \\
(Crop) 6\%               & 0.52 & 0.47 & 0.65 \\
(Salt \& Pepper) 0.05    & 0.55 & 0.89 & 0.90 \\
(Salt \& Pepper) 0.10    & 0.40 & 0.82 & 0.88 \\
(Perspective) \(D=0.7, p=0.7\) & 0.74 & 0.89 & 0.90 \\
(Perspective) \(D=1.0, p=1.0\) & 0.29 & 0.45 & 0.68 \\ \hline
\end{tabular}
\vspace{0.5em} 
\caption{Cosine Similarity Comparison on OXFORD-IIT PET}
\label{tab:oxford_compare}
\vspace{-2.5em}
\end{table}

\begin{table}[h]
\centering
\renewcommand{\arraystretch}{1.2} 
\begin{tabular}{lccc} 
\hline
\textbf{Distortion} & \textbf{SimCLR} & \textbf{DINO} & \textbf{Our Model} \\ \hline
(Rotation) \(30^\circ\)   & 0.75 & 0.90 & 0.94 \\
(Rotation) \(70^\circ\)   & 0.58 & 0.83 & 0.92 \\
(Crop) 2\%               & 0.48 & 0.37 & 0.52 \\
(Crop) 6\%               & 0.48 & 0.35 & 0.50 \\
(Salt \& Pepper) 0.05    & 0.73 & 0.87 & 0.92 \\
(Salt \& Pepper) 0.10    & 0.60 & 0.79 & 0.90 \\
(Perspective) \(D=0.7, p=0.7\) & 0.74 & 0.83 & 0.89 \\
(Perspective) \(D=1.0, p=1.0\) & 0.29 & 0.45 & 0.64 \\ \hline
\end{tabular}
\vspace{0.5em} 
\caption{Cosine Similarity Comparison on COCO}

\label{tab:coco_compare}
\vspace{-2.0em}
\end{table}

The comparative results demonstrate that our model consistently outperforms both SimCLR and DINO across all distortion conditions. Notably, our approach maintains higher cosine similarity values, even under severe transformations such as strong perspective distortions and high levels of salt-and-pepper noise. This indicates that our text-guided invariant learning method is more robust in preserving feature integrity, ensuring that the learned representations remain stable despite substantial image alterations. Compared to DINO, which has been explored in other watermarking schemes, our model achieves greater invariance, particularly under extreme distortions, highlighting its effectiveness for robust image watermarking applications.

To further demonstrate that the proposed feature space is not only robust but also retains semantic information, we conducted a linear evaluation experiment on an unseen dataset. For this purpose, we appended a classification layer to each of the compared feature extractor models and trained these layers on the CIFAR-10 training set while keeping the backbone frozen. The CIFAR-10 test set, consisting of 10,000 images resized to \(224 \times 224\) pixels, was used to evaluate performance under various distortions. 
We applied a range of distortions at different intensities, including rotations (\(10^\circ\) and \(20^\circ\)), brightness adjustments (factors of 0.3 and 0.5), contrast modifications (factors of 0.3 and 0.5), and Gaussian blur (kernel sizes of 1.0 and 3.0). Classification accuracy (in percentage) was recorded for each distortion condition.
\vspace{-0.75em}
\begin{table}[htbp]
    \centering
    \footnotesize 
    \renewcommand{\arraystretch}{1.2} 
    
    \resizebox{\linewidth}{!}{%
    \begin{tabular}{lcccccc}
    \hline
    \textbf{Distortions} & \textbf{VIC Reg} & \textbf{BYOL} & \textbf{SimCLR} & \textbf{Dino} & \textbf{Our Model} \\
    \hline
    None           & 69.72  & 83.93 & 61.81 & 91.90 & 92.48 \\
    (Rot.) 10     & 54.41 & 67.49 & 50.78 & 84.86 & 89.37 \\
    (Rot.) 20     & 38.37 & 48.79 & 35.03 & 72.85 & 85.52 \\
    (Bright.) 0.3  & 66.68 & 83.31 & 60.70 & 91.61 & 91.67 \\
    (Bright.) 0.5 & 68.73  & 81.69 & 58.51 & 91.27 & 90.94 \\
    (Cont.) 0.3   & 69.32  & 83.67 & 60.95 & 91.77 & 91.91 \\
    (Cont.) 0.5   & 67.98  & 82.54 & 59.86 & 91.63 & 91.09 \\
    Blur(1.0)      & 61.80  & 76.13 & 40.00 & 91.90 & 90.44 \\
    Blur(3.0)      & 55.21  & 68.83 & 36.75 & 91.90 & 89.63 \\
    \hline
    \end{tabular}%
    }
    \vspace{0.25em}
    \caption{Linear Evaluation on Unseen CIFAR-10 Dataset With Different Distortions (\%)}
    \label{tab:linear_evaluation}
\end{table}
\vspace{-1.0em}

\vspace{-1.0em}

The results in Table~\ref{tab:linear_evaluation} show that our model consistently outperforms state-of-the-art methods, including SimCLR, BYOL~\cite{grill2020bootstrap}, DINO, and VICReg~\cite{vicreg2021}, across all distortions. This highlights the effectiveness of our text-guided invariant learning approach in capturing both robust and semantically meaningful features, ensuring high classification accuracy even under challenging transformations.

\vspace{-1.0em}
\subsection{Robust Image Watermarking with Invariant Feature}
\label{robust_watermarking}
\vspace{-0.25em}

In this section, we conduct a multi-bit watermarking analysis under various distortions commonly encountered in image processing, comparing the performance of our model with that of~\cite{fernandez2022watermarking}, which utilizes a pretrained DINO model as a feature extractor. For this experiment, we use a subset of the Oxford 102 Flowers dataset, selected for its diverse collection of high-quality images, making it well-suited for evaluating image manipulation techniques. All images are resized to \(224 \times 224\) pixels to ensure consistency across experiments.

\begin{figure}[ht]
    \centering
    \begin{minipage}{0.2\linewidth}
        \centering
        \includegraphics[width=\linewidth]{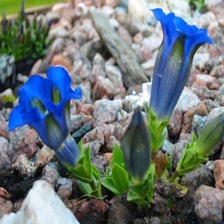}\\
        \footnotesize (a) Original
    \end{minipage}\hspace{0.02\linewidth}%
    \begin{minipage}{0.2\linewidth}
        \centering
        \includegraphics[width=\linewidth]{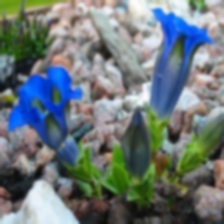}\\
        \footnotesize (b) Blur
    \end{minipage}\hspace{0.02\linewidth}%
    \begin{minipage}{0.2\linewidth}
        \centering
        \includegraphics[width=\linewidth]{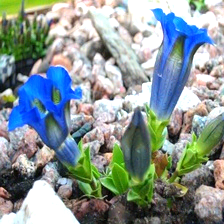}\\
        \footnotesize (c) Brightness
    \end{minipage}\hspace{0.02\linewidth}%
    \\
    \begin{minipage}{0.2\linewidth}
        \centering
        \includegraphics[width=\linewidth]{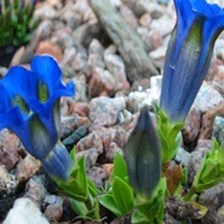}\\
        \footnotesize (d) Crop
    \end{minipage}\hspace{0.02\linewidth}%
    \begin{minipage}{0.2\linewidth}
        \centering
        \includegraphics[width=\linewidth]{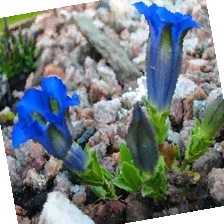}\\
        \footnotesize (e) Rotation
    \end{minipage}\hspace{0.02\linewidth}%
    \begin{minipage}{0.2\linewidth}
        \centering
        \includegraphics[width=\linewidth]{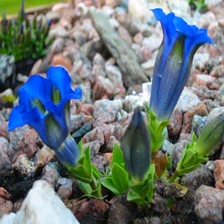}\\
        \footnotesize (f) Saturation
    \end{minipage}
    \caption{\footnotesize Distortions used in our watermarking analysis}
    \label{fig:dis_example}
\end{figure}

A 10-bit watermark message is embedded while maintaining a peak signal-to-noise ratio (PSNR) of 40. To ensure fair comparison, we generate a random key with dimensions matching the output feature space of both our model and~\cite{fernandez2022watermarking}. The watermarking process follows the methodology outlined in Section~\ref{sec: wm_method}. Both models are trained for 100 epochs to embed watermarks into the images, and the resulting watermarked images are saved for subsequent extraction experiments.

\begin{table}[htbp]
\vspace{-0.75em}
    \centering
    \renewcommand{\arraystretch}{1.2} 
    \begin{tabular}{lcc}
    \hline
    \textbf{Distortion} & \textbf{\cite{fernandez2022watermarking}} & \textbf{Our Model} \\
    \hline
    (Blur) k=3        & 0.85 & 0.93 \\
    (Blur) k=7        & 0.73 & 0.87 \\
    (Blur) k=9        & 0.63 & 0.76 \\
    (Crop) 5\%        & 0.82 & 0.84 \\
    (Crop) 10\%       & 0.78 & 0.81 \\
    (Crop) 15\%       & 0.74 & 0.81 \\
    (Rotation) 2      & 0.83 & 0.84 \\
    (Rotation) 6      & 0.76 & 0.76 \\
    (Rotation) 10     & 0.62 & 0.67 \\
    (Bright.) 1.0     & 0.95 & 0.95 \\
    (Bright.) 1.1     & 0.94 & 0.94 \\
    (Bright.) 1.2     & 0.93 & 0.93 \\
    (Saturation) 1.0  & 0.95 & 0.95 \\
    (Saturation) 1.05 & 0.95 & 0.95 \\
    (Saturation) 1.1  & 0.95 & 0.95 \\
    \hline
    \end{tabular}
    \vspace{0.5em}
    \caption{Watermark Extraction Bit Accuracy under Various Distortions and Comparison with Dino}
    \label{tab:watermark_results}
    \vspace{-2.5em}
\end{table}

For robust watermark extraction, we evaluate the robustness of the embedded watermark under various attack scenarios, such as different levels of blur, cropping, rotation, brightness, and saturation distortions (see examples in Fig.~\ref{fig:dis_example}). As summarized in Table~\ref{tab:watermark_results}, our model consistently achieves higher extraction accuracy, particularly under severe distortions such as strong blur, cropping, and rotation, where it significantly outperforms the DINO-based approach. Under milder distortions, such as brightness and saturation adjustments, both models exhibit comparable performance. These results highlight the effectiveness of our method in preserving watermark integrity under challenging conditions, demonstrating superior robustness compared to state-of-the-art self-supervised feature extractor based image watermarking.

%% file: section/conclusion.tex
This paper introduces a novel text-guided invariant feature learning framework for robust image watermarking, leveraging text embeddings as semantic anchors to enforce feature consistency under distortions. Unlike existing self-supervised learning approaches that passively inherit invariance, our method explicitly optimizes for robustness, ensuring stable feature representations. Experimental results demonstrate that our model outperforms state-of-the-art SSL methods in maintaining feature integrity and achieving higher watermark extraction accuracy under severe distortions. These findings highlight the effectiveness of text-guided contrastive learning in enhancing watermarking robustness.

%% file: main.bbl
\begin{thebibliography}{10}
\providecommand{\url}[1]{#1}
\csname url@samestyle\endcsname
\providecommand{\newblock}{\relax}
\providecommand{\bibinfo}[2]{#2}
\providecommand{\BIBentrySTDinterwordspacing}{\spaceskip=0pt\relax}
\providecommand{\BIBentryALTinterwordstretchfactor}{4}
\providecommand{\BIBentryALTinterwordspacing}{\spaceskip=\fontdimen2\font plus
\BIBentryALTinterwordstretchfactor\fontdimen3\font minus \fontdimen4\font\relax}
\providecommand{\BIBforeignlanguage}[2]{{%
\expandafter\ifx\csname l@#1\endcsname\relax
\typeout{** WARNING: IEEEtran.bst: No hyphenation pattern has been}%
\typeout{** loaded for the language `#1'. Using the pattern for}%
\typeout{** the default language instead.}%
\else
\language=\csname l@#1\endcsname
\fi
#2}}
\providecommand{\BIBdecl}{\relax}
\BIBdecl

\bibitem{zhong2023brief}
X.~Zhong, A.~Das, F.~Alrasheedi, and A.~Tanvir, ``A brief, in-depth survey of deep learning-based image watermarking,'' \emph{Applied Sciences}, vol.~13, no.~21, p. 11852, 2023.

\bibitem{karki2024deep}
B.~Karki, C.-H. Tsai, P.-C. Huang, and X.~Zhong, ``Deep learning-based text-in-image watermarking,'' in \emph{2024 IEEE 7th International Conference on Multimedia Information Processing and Retrieval (MIPR)}.\hskip 1em plus 0.5em minus 0.4em\relax IEEE, 2024, pp. 376--382.

\bibitem{fernandez2022watermarking}
P.~Fernandez, A.~Sablayrolles, T.~Furon, H.~J{\'e}gou, and M.~Douze, ``Watermarking images in self-supervised latent spaces,'' in \emph{ICASSP 2022-2022 IEEE International Conference on Acoustics, Speech and Signal Processing (ICASSP)}.\hskip 1em plus 0.5em minus 0.4em\relax IEEE, 2022, pp. 3054--3058.

\bibitem{lecun1998cnn}
Y.~LeCun, L.~Bottou, Y.~Bengio, and P.~Haffner, ``Gradient-based learning applied to document recognition,'' \emph{Proceedings of the IEEE}, vol.~86, no.~11, pp. 2278--2324, 1998.

\bibitem{autoaugment2018}
E.~D. Cubuk, B.~Zoph, D.~Mane, V.~Vasudevan, and Q.~V. Le, ``Autoaugment: Learning augmentation strategies from data,'' in \emph{Proceedings of the IEEE/CVF conference on computer vision and pattern recognition}, 2019, pp. 113--123.

\bibitem{zhang2021learning}
H.~Zhang and T.~Arodz, ``Learning invariance in deep neural networks,'' in \emph{International Conference on Computational Science}.\hskip 1em plus 0.5em minus 0.4em\relax Springer, 2021, pp. 64--74.

\bibitem{immer2021invariance}
A.~Immer, T.~van~der Ouderaa, G.~R{\"a}tsch, V.~Fortuin, and M.~van~der Wilk, ``Invariance learning in deep neural networks with differentiable laplace approximations,'' \emph{Advances in Neural Information Processing Systems}, vol.~35, pp. 12\,449--12\,463, 2022.

\bibitem{laplace1992}
D.~J. MacKay, ``A practical bayesian framework for backpropagation networks,'' \emph{Neural computation}, vol.~4, no.~3, pp. 448--472, 1992.

\bibitem{kfac2015}
J.~Martens and R.~Grosse, ``Optimizing neural networks with kronecker-factored approximate curvature,'' in \emph{International conference on machine learning}.\hskip 1em plus 0.5em minus 0.4em\relax PMLR, 2015, pp. 2408--2417.

\bibitem{chen2020simple}
T.~Chen, S.~Kornblith, M.~Norouzi, and G.~Hinton, ``A simple framework for contrastive learning of visual representations,'' in \emph{International conference on machine learning}.\hskip 1em plus 0.5em minus 0.4em\relax PMLR, 2020, pp. 1597--1607.

\bibitem{grill2020bootstrap}
J.-B. Grill, F.~Strub, F.~Altch{\'e}, C.~Tallec, P.~Richemond, E.~Buchatskaya, C.~Doersch, B.~Avila~Pires, Z.~Guo, M.~Gheshlaghi~Azar \emph{et~al.}, ``Bootstrap your own latent-a new approach to self-supervised learning,'' \emph{Advances in neural information processing systems}, vol.~33, pp. 21\,271--21\,284, 2020.

\bibitem{caron2021emerging}
M.~Caron, H.~Touvron, I.~Misra, H.~J{\'e}gou, J.~Mairal, P.~Bojanowski, and A.~Joulin, ``Emerging properties in self-supervised vision transformers,'' in \emph{Proceedings of the IEEE/CVF international conference on computer vision}, 2021, pp. 9650--9660.

\bibitem{yao2024empowering}
T.~Yao, Y.~Chen, Z.~Chen, K.~Hu, Z.~Shen, and K.~Zhang, ``Empowering graph invariance learning with deep spurious infomax,'' \emph{arXiv preprint arXiv:2407.11083}, 2024.

\bibitem{vukotic2020classification}
V.~Vukoti{\'c}, V.~Chappelier, and T.~Furon, ``Are classification deep neural networks good for blind image watermarking?'' \emph{Entropy}, vol.~22, no.~2, p. 198, 2020.

\bibitem{radford2021learning}
A.~Radford, J.~W. Kim, C.~Hallacy, A.~Ramesh, G.~Goh, S.~Agarwal, G.~Sastry, A.~Askell, P.~Mishkin, J.~Clark \emph{et~al.}, ``Learning transferable visual models from natural language supervision,'' in \emph{International conference on machine learning}.\hskip 1em plus 0.5em minus 0.4em\relax PmLR, 2021, pp. 8748--8763.

\bibitem{hodosh2013framing}
M.~Hodosh, P.~Young, and J.~Hockenmaier, ``Framing image description as a ranking task: Data, models and evaluation metrics,'' \emph{Journal of Artificial Intelligence Research}, vol.~47, pp. 853--899, 2013.

\bibitem{plummer2015flickr30k}
B.~Plummer, L.~Wang, C.~Cervantes, J.~Caicedo, J.~Hockenmaier, and S.~Lazebnik, ``Flickr30k entities: Collecting region-to-phrase correspondences for richer image understanding,'' in \emph{Proceedings of the IEEE International Conference on Computer Vision (ICCV)}, 2015.

\bibitem{parkhi2012cats}
O.~M. Parkhi, A.~Vedaldi, and A.~Zisserman, ``Cats and dogs,'' in \emph{British Machine Vision Conference (BMVC)}, 2012.

\bibitem{article}
A.~Coates, A.~Ng, and H.~Lee, ``An analysis of single-layer networks in unsupervised feature learning,'' \emph{Journal of Machine Learning Research}, vol.~15, pp. 215--223, 01 2011.

\bibitem{oquab2024dinov2}
M.~Oquab, T.~Darcet, T.~Moutakanni, H.~Vo, M.~Szafraniec, V.~Khalidov, P.~Fernandez, D.~Haziza, F.~Massa, A.~El-Nouby \emph{et~al.}, ``Dinov2: Learning robust visual features without supervision,'' \emph{Transactions on Machine Learning Research Journal}, pp. 1--31, 2024.

\bibitem{lin2014microsoft}
T.-Y. Lin, M.~Maire, S.~Belongie, L.~Bourdev, R.~Girshick, J.~Hays, P.~Perona, D.~Ramanan, C.~L. Zitnick, and P.~Doll{\'a}r, ``Microsoft coco: Common objects in context,'' in \emph{European Conference on Computer Vision (ECCV)}, 2014, pp. 740--755.

\bibitem{vicreg2021}
A.~Bardes, J.~Ponce, and Y.~LeCun, ``Vicreg: Variance-invariance-covariance regularization for self-supervised learning,'' 2021.

\end{thebibliography}
